\documentclass{article}

\usepackage[eandd, nonanonymous,final]{neurips_2026}

\usepackage[utf8]{inputenc} 
\usepackage[T1]{fontenc}    
\usepackage{hyperref}       
\usepackage{url}            
\usepackage{booktabs}       
\usepackage{amsfonts}       
\usepackage{nicefrac}       
\usepackage{microtype}      
\usepackage{xcolor}         
\usepackage{graphicx}
\usepackage{adjustbox}
\usepackage{amsmath}
\usepackage{float}
\usepackage{comment}

\definecolor{grank1}{RGB}{0,60,15}
\definecolor{grank2}{RGB}{15,75,30}
\definecolor{grank3}{RGB}{30,90,45}
\definecolor{grank4}{RGB}{45,105,60}
\definecolor{grank5}{RGB}{60,120,75}
\definecolor{grank6}{RGB}{90,150,105}
\definecolor{grank7}{RGB}{120,180,135}
\definecolor{grank8}{RGB}{150,200,165}
\definecolor{grank9}{RGB}{180,215,195}
\definecolor{grank10}{RGB}{205,225,215}
\newcommand{\gcolortext}[2]{\textcolor{grank#1}{#2}}
\definecolor{rrank1}{RGB}{120,0,20}
\definecolor{rrank2}{RGB}{140,20,40}
\definecolor{rrank3}{RGB}{160,40,60}
\definecolor{rrank4}{RGB}{180,60,80}
\definecolor{rrank5}{RGB}{200,80,100}
\definecolor{rrank6}{RGB}{220,100,120}
\definecolor{rrank7}{RGB}{240,140,160}
\definecolor{rrank8}{RGB}{250,170,190}
\definecolor{rrank9}{RGB}{255,195,210}
\definecolor{rrank10}{RGB}{255,220,230}
\newcommand{\rcolortext}[2]{\textcolor{rrank#1}{#2}}
\definecolor{brank1}{RGB}{0,20,90}
\definecolor{brank2}{RGB}{15,35,110}
\definecolor{brank3}{RGB}{30,50,130}
\definecolor{brank4}{RGB}{45,65,150}
\definecolor{brank5}{RGB}{60,80,170}
\definecolor{brank6}{RGB}{95,115,200}
\definecolor{brank7}{RGB}{135,150,220}
\definecolor{brank8}{RGB}{170,180,235}
\definecolor{brank9}{RGB}{200,210,245}
\definecolor{brank10}{RGB}{220,225,250}
\newcommand{\bcolortext}[2]{\textcolor{brank#1}{#2}}
\definecolor{yrank1}{RGB}{102,51,0}
\definecolor{yrank2}{RGB}{128,64,0}
\definecolor{yrank3}{RGB}{153,76,0}
\definecolor{yrank4}{RGB}{179,89,0}
\definecolor{yrank5}{RGB}{204,102,0}
\definecolor{yrank6}{RGB}{220,140,40}
\definecolor{yrank7}{RGB}{230,170,90}
\definecolor{yrank8}{RGB}{240,200,140}
\definecolor{yrank9}{RGB}{245,220,180}
\definecolor{yrank10}{RGB}{250,235,210}
\newcommand{\ycolortext}[2]{\textcolor{yrank#1}{#2}}
\definecolor{prank1}{RGB}{0,0,0}
\definecolor{prank2}{RGB}{30,30,30}
\definecolor{prank3}{RGB}{60,60,60}
\definecolor{prank4}{RGB}{85,85,85}
\definecolor{prank5}{RGB}{110,110,110}
\definecolor{prank6}{RGB}{140,140,140}
\definecolor{prank7}{RGB}{165,165,165}
\definecolor{prank8}{RGB}{190,190,190}
\definecolor{prank9}{RGB}{210,210,210}
\definecolor{prank10}{RGB}{225,225,225}
\newcommand{\pcolortext}[2]{\textcolor{prank#1}{#2}}

\title{CTF4Nuclear:  Common Task Framework for Nuclear Fission and Fusion Models}

\author{%
  Stefano Riva$^{1}$,\And
  Carolina Introini$^{2}$,\And
  Antonio Cammi$^{2}$,\And
  Dean Price$^{3}$, \And
  Alexey Yermakov$^{4,5}$,\And
  Yue Zhao$^{6}$,\And
  Philippe M. Wyder$^{7}$,\And
  Judah Goldfeder$^{8}$,\And
  Jan Williams$^{9}$,\And
  Amy Sara Rude$^{4}$,\And
  Matteo Tomasetto$^{10}$,\And
  Joe Germany$^{11}$,\And
  Joseph Bakarji$^{12}$,\And
  Georg Maierhofer$^{13}$,\And
  Miles Cranmer$^{13}$,\And
  J. Nathan Kutz$^{1,4,5}$
  \thanks{Corresponding author: \texttt{kutz@uw.edu}} \AND
  \\
  $^{1}$Autodesk Research, London, UK\\
  $^{2}$Department of Energy, Nuclear Engineering Division, Politecnico di Milano, Milan, Italy \\
  $^{3}$Nuclear Science and Engineering, Massachusetts Institute of Technology, Cambridge, MA 02139 \\
  $^{4}$Department of Applied Mathematics, University of Washington, Seattle, WA 98195 \\
  $^{5}$Department of Electrical and Computer Engineering, University of Washington, Seattle, WA 98195 \\
  $^{6}$High Performance Machine Learning, SURF, Amsterdam, the Netherlands \\
  $^{7}$Distyl AI, New York, NY 10016 \\
  $^{8}$Department of Computer Science, Columbia University, New York, NY 10027 \\
  $^{9}$Department of Mechanical Engineering, University of Washington, Seattle, WA 98195 \\
  $^{10}$Department of Mechanical Engineering, Politecnico di Milano, Milan, Italy \\
  $^{11}$Department of Mathematics, American University in Beirut, Beirut, Lebanon \\
  $^{12}$Department of Mechanical Engineering, American University in Beirut, Beirut, Lebanon \\
  $^{13}$Department of Applied Mathematics and Theoretical Physics, University of Cambridge, Cambridge, UK\\
}

\begin{document}

\maketitle

\begin{abstract}
    The demand for clean energy is ever increasing, with new nuclear technologies presenting a complementary solution to renewable energies. However, designing and operating these systems is exceptionally difficult, given the complexity of the physical phenomena that interact to form the system dynamics. While high-fidelity simulations help to understand the non-linear, multi-physics interactions within a reactor (e.g., neutronics and thermal-hydraulics), they are computationally expensive and rarely suitable for real-time applications. Furthermore, model-based approaches are inherently sensitive to simplifying assumptions required to derive their governing equations and parameters, leading to inevitable discrepancies with real-world measurements. In contrast, Machine Learning (ML) methods have the potential to generate reliable surrogate models which may be able to quickly predict the system's behaviour. However, the number of data-driven methods that can potentially be used for this task is extremely large and diverse. In a safety-critical setting such as nuclear engineering, a fair comparison of different ML methods, and a clear understanding of their advantages and limitations, is thus of paramount importance. To address this, we introduce a Common Task Framework (CTF) for the application of ML in nuclear engineering, building upon previous efforts in dynamical systems and seismology. This CTF considers a curated set of datasets from different nuclear and nuclear-adjacent systems (e.g., molten salt, micro-reactors, electrically conducting fusion coolants, real experimental facilities). The CTF rigorously evaluates the performance of a method on twelve established metrics (forecasting, noise robustness, limited data, and parametric generalisation), alongside a new paradigm focused on system monitoring from sparse measurements only. We illustrate the framework by benchmarking standard ML baselines against these datasets, revealing current method limitations. Our vision is to replace {\em ad hoc} comparisons with standardized evaluations on hidden test sets, raising the bar for rigour and reproducibility in scientific ML for the nuclear industry.
\end{abstract}

\section{Introduction}
\label{sec:intro}

Nuclear fission and fusion reactors are among the most safety-critical dynamical systems ever engineered, requiring continuous real-time knowledge of the full physical state of the reactor core: neutron flux, temperature, coolant velocity, and delayed-neutron precursor concentrations \cite{DuderstadtHamilton}. These requirements are becoming more stringent as nuclear plants are expected to operate in load-following mode alongside variable renewable sources \cite{international2023iaea, RUTH2014684}, and as next-generation digital twins \cite{Grieves_DT, mohanty_physics-infused_2021, iaea_2025_ai_report} are required to provide real-time state estimates for control, monitoring, and accident prognosis. Nuclear cores are extreme environments, with in-core instrumentation becoming almost physically impossible in Generation-IV or fusion concepts, forcing sensors to the core boundary or external shielding structures \cite{ICAPP_plus2023, PHYSOR24_MSFR_outcore}. Recovering the full multi-physics state from these sparse, out-of-core observations is an inherently ill-conditioned inverse problem \cite{argaud_sensor_2018} that involves fields that the sensors never directly measure. The governing equations, coupling the neutron Boltzmann transport equation with the Navier-Stokes system \cite{DuderstadtHamilton, demaziere_6_2020}, yield Full Order Models (FOMs) whose runtime spans hours to days, preventing real-time use \cite{Kapteyn_Willcox_2020}.

As a result of the complexity of nuclear FOMs, the nuclear engineering community is developing a growing number of surrogate and data-driven modelling approaches. Reduced Order Modelling (ROM) based on the Proper Orthogonal Decomposition (POD) \cite{lassila_model_2014, rozza_model_2020} provides a core compression framework, with projection-based approaches such as GeN-ROM \cite{GERMAN2022104148} extending this to fully intrusive multi-physics surrogates. Non-intrusive methods that integrate measurement data include the Generalised Empirical Interpolation Method \cite{maday_generalized_2015} and the Parametrized-Background Data-Weak formulation \cite{maday_parameterized-background_2014}, which are applied to nuclear state estimation, sensor placement, and multi-physics bias correction \cite{gong_generalized_2022, ICAPP_plus2023, gong_parameter_2023}. More recently, deep learning methods have entered the nuclear domain. Physics-Informed Neural Networks \cite{raissi2019physics} have been applied to point-kinetics transients \cite{prantikos2022pinn, prantikos2023tlpinn}, neutron diffusion \cite{wang2022surrogate}, and full transport modelling \cite{geng2026pinn}. Neural operators have emerged as powerful function-space surrogates: Deep Operator Networks \cite{lu2021deeponet} have been used for particle transport and thermal-hydraulic monitoring \cite{kobayashi2024deeponet, alam2024virtual}, while Fourier Neural Operators \cite{li2021fno} have demonstrated six-orders-of-magnitude computational acceleration over traditional solvers in predicting plasma dynamics within Tokamak reactors \cite{gopakumar2024fno}, making them a natural candidate for fusion applications also included in our benchmark. Sequence-to-state architectures such as Shallow Recurrent Decoders \cite{williams_sensing_2023, kutz_shallow_2024} reconstruct the full core state from as few as three out-of-core sensors, with strong performance on molten salt reactor benchmarks including parametric transients \cite{riva2024robuststateestimationpartial, shredrom}. Broader reviews of the field are available in \cite{zhang2025aireactor} and \cite{huang2023aireview}.

Despite this rapid growth, the field faces a reproducibility crisis documented across scientific machine learning more broadly \cite{McGreivy2024}: methods are evaluated on self-selected datasets, with both training and test data available to developers, creating structural incentives for optimistic reporting. Without a truly withheld test set, guarding against selective re-training until a favourable outcome is obtained is impossible. Weak baselines, inconsistent protocols, and non-reproducible results follow predictably. The nuclear domain amplifies this problem: experimental data from operating reactors are scarce and proprietary; full-scale demonstrators for Generation-IV concepts do not exist; and high-fidelity simulation datasets, the de facto reference, are rarely shared in standardised formats. The OECD/NEA has acknowledged the need through its Task Force on AI/ML for Scientific Computing in Nuclear Engineering \cite{nea2024taskforce}, with benchmark exercises on critical heat flux and reactor kinetics; yet these efforts remain institution-specific, lack a community-accessible withheld test set, and do not cover the transient state-estimation and surrogate-modelling tasks most critical for digital twin applications.

\begin{figure}[t]
  \centering
  \includegraphics[width=1\linewidth]{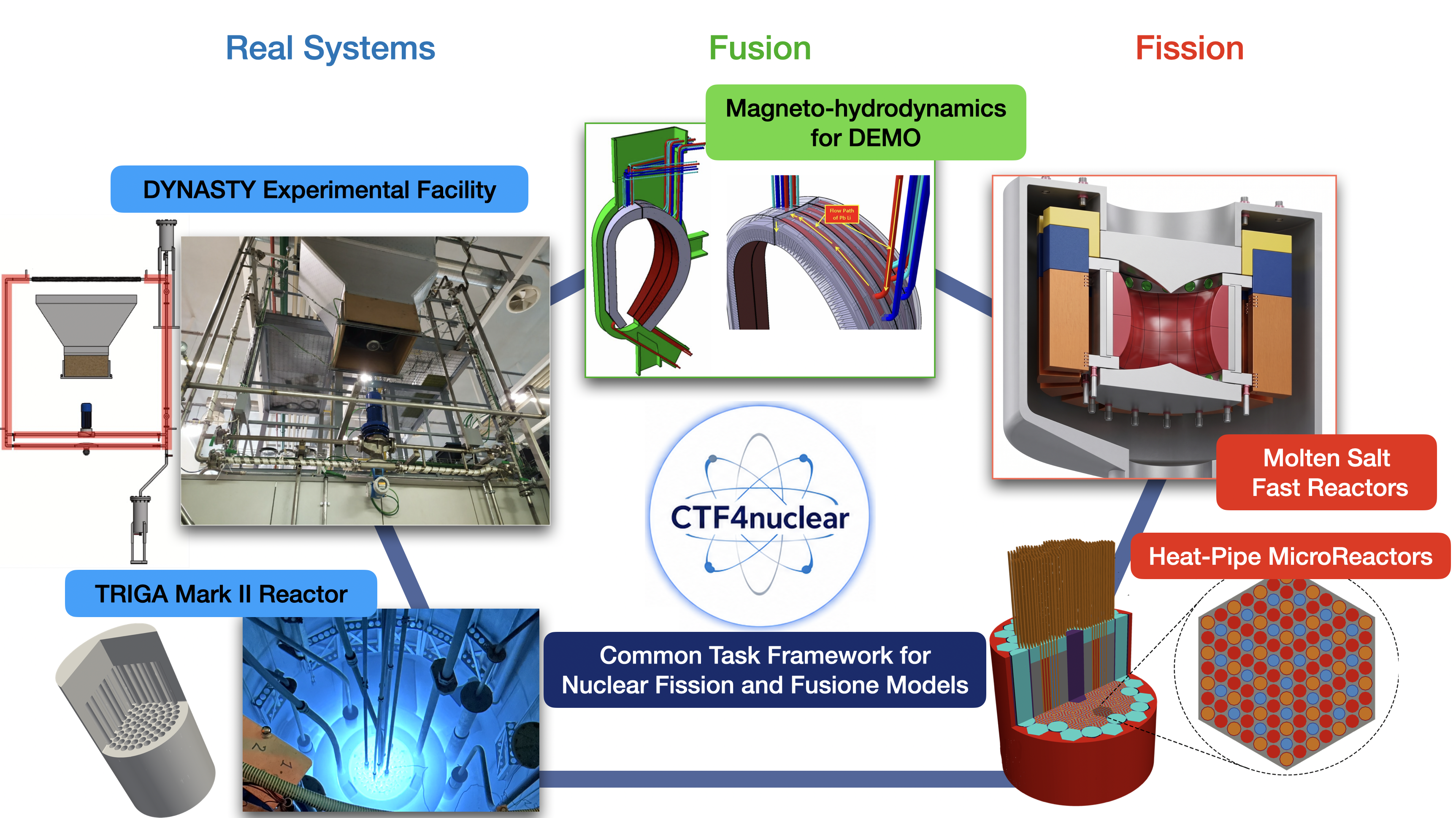}
  \caption{Scheme of the Common Task Framework (CTF) for Nuclear Fission and Fusion Models, with different datasets: Molten Salt Fast Reactor \cite{brovchenko2013optimization}, Heat-Pipe MicroReactor \cite{PRINCE2024110365}, Magneto-Hydrodynamics (MHD) case studies \cite{magnetohdfoam2024} and real experimental data from DYNASTY \cite{pini_experimental_2016} and TRIGA \cite{international2016iaeaTRIGA}.}
  \label{fig:ctf_overview}
\end{figure}

\subsection{Related Work}

Common Task Frameworks (CTFs) have been the primary mechanism by which ML communities establish rigorous, reproducible progress metrics \cite{donoho2017}. The transformative role of landmark CTFs is well-established: ImageNet \cite{deng2009imagenet} catalysed deep convolutional networks \cite{krizhevsky2012imagenet}; competitive games enabled AlphaZero \cite{silver2018general}; the Arcade Learning Environment \cite{bellemare2013} and OpenAI Gym \cite{brockman2016openaigym} accelerated reinforcement learning; CASP \cite{jumper2021alphafold} drove AlphaFold in structural biology; and natural language processing has benefited from code-generation \cite{chen2021codex} and mathematical reasoning benchmarks \cite{hendrycks2021math, cobbe2021gsm8k}, with recent systems such as DeepSeek-R1 using these to demonstrate competitive performance \cite{deepseek2025r1}.

CTFs for scientific ML have recently gained momentum. Wyder et al. \cite{wyder2025commontaskframeworkcritical} proposed a framework for dynamical systems, benchmarking methods on the Kuramoto-Sivashinsky PDE and the Lorenz attractor across twelve task-specific metrics covering forecasting, reconstruction, noise, limited data, and parametric generalisation. Yermakov et al. \cite{yermakov2025seismicwavefieldcommontask} extended this to seismological wavefield modelling with three datasets at global, crustal, and local scales. Other scientific benchmarking efforts include PDEBench \cite{takamoto2022pdebench}, The Well \cite{ohana2024well}, and CoDBench \cite{burark2024codbench}, which provide datasets and evaluation protocols for learning to solve differential equations, but which allow for self-reporting. None of these targets nuclear reactor systems (both fission and fusion) whose physics, observability constraints, and safety requirements are sufficiently distinct to warrant a dedicated framework.

\subsection{Common Task Framework for Nuclear Reactors}

We introduce \textsc{CTF4Nuclear}, the first open common task framework (benchmarking) platform for ML surrogate modelling and state estimation of nuclear reactor systems, built on the \texttt{ctf4science} package \cite{wyder2025commontaskframeworkcritical, yermakov2025seismicwavefieldcommontask} and openly available at our \href{https://github.com/CTF-for-Science/ctf4science}{GitHub repository}.

A distinguishing feature of \textsc{CTF4Nuclear} is the deliberate diversity of its benchmark suite. We collect datasets from five physically distinct nuclear engineering systems (see Fig. \ref{fig:ctf_overview}): the Molten Salt Fast Reactor (MSFR) \cite{brovchenko2013optimization, aufiero2014development}, a Generation-IV circulating-fuel concept with fully coupled neutronics and thermal-hydraulics and strictly out-of-core sensors; a Heat-Pipe MicroReactor (HPMR) \cite{PRINCE2024110365, wang2025griffin}, a compact solid-fuel concept with a different observability structure; electrically conducting coolants under magneto-hydrodynamic effects \cite{magnetohdfoam2024} for fusion energy applications; and real experimental facilities including the TRIGA Mark II research reactor \cite{international2016iaeaTRIGA} and the DYNASTY natural-circulation loop \cite{pini_experimental_2016}, which introduce genuine measurement noise and model-reality mismatch absent from purely synthetic datasets. This breadth is intentional: a method that succeeds on one system but fails on others reveals scope limitations invisible in single-dataset comparisons.

In addition to the twelve task-specific metrics of \cite{wyder2025commontaskframeworkcritical}, covering forecasting and state reconstruction under noise, limited data, and parametric variability, \textsc{CTF4Nuclear} introduces a task class specific to nuclear monitoring: the reconstruction of physically unobservable fields (such as delayed neutron precursor concentrations or core velocity) from sparse measurements. This task has no analogue in existing scientific CTFs and directly targets the central challenge described above. \\
A composite score is computed as the mean of all task scores; the multi-metric design prevents winner-takes-all outcomes and exposes method profiles (a method robust to noise may perform poorly under parametric shift, for instance), helping practitioners select methods appropriate to their deployment context. A truly withheld test set, evaluated by an independent referee and published on a Kaggle leaderboard, ensures that no participant can overfit to the evaluation set.

\section{Datasets \& Evaluation Metrics} 
The CTF4Nuclear extends the previous CTFs for scientific ML \cite{wyder2025commontaskframeworkcritical, yermakov2025seismicwavefieldcommontask} by introducing a new set of datasets related to nuclear fission and fusion. The datasets are selected to cover a wide range of systems, from water-based technology such as the TRIGA Mark II reactor \cite{international2016iaeaTRIGA} and the DYNASTY experimental facility \cite{pini_experimental_2016}, towards more advanced fission reactor technologies, like  Heat Pipe Microreactor \cite{PRINCE2024110365} and molten salt reactors \cite{brovchenko2013optimization}, and magneto-hydrodynamics for fusion applications \cite{ferrero2024impact}. Figure \ref{fig:ctf_overview} shows some pictures of the geometries of the datasets considered; we stress that due to the modularity of the CTF package it allows to include new reactors/systems in a rather simple way.

\subsection{Dataset details}
\begin{figure}[t]
    \centering
    \includegraphics[width=0.9\linewidth]{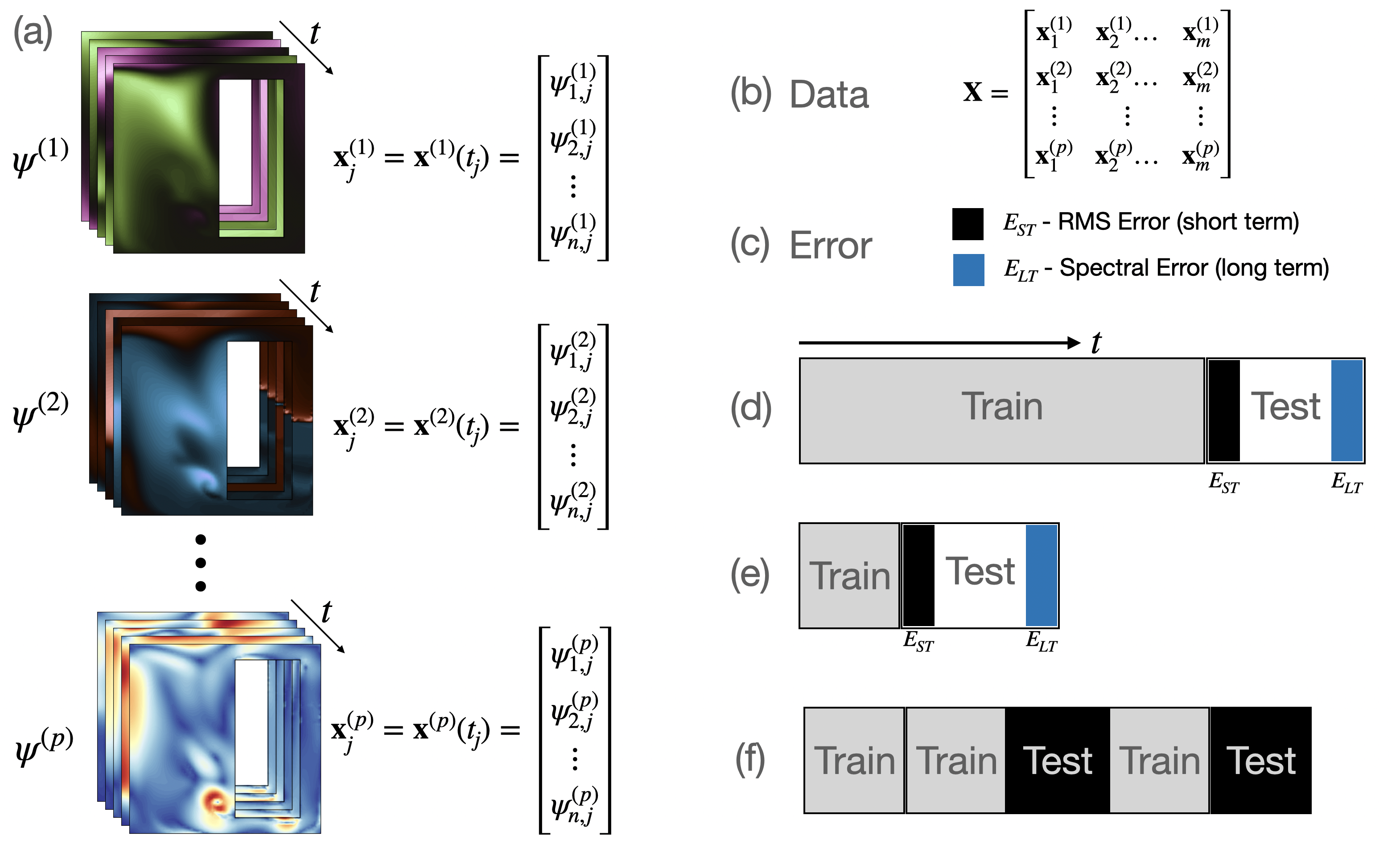}
    \caption{ The CTF scores the performance of methods on nuclear reactors data. (a) Data is collected and organized into matrices, which is then split into testing and training sets. (b) Multiple fields $(\psi_1, \psi_2, ... \psi_p)$ are vertically concatenated. (c) RMSE errors are computed for reconstruction and short-time forecasting, while the spectral error computes the statistics of long-time forecasting (spatial or temporal). (d) Forecasting and reconstruction tasks are evaluated on noise-free, low-noise and high-noise data. Methods are also evaluated when (e) Only limited data is available and (f) For reconstruction of parametrically dependent data.}
    \label{fig: datametrics}
\end{figure}

Figure \ref{fig: datametrics}a shows the data collection when multiple fields are involved: for any field $\psi^{(i)}$, the state at time $t_j$, $j=1, \dots, m$, corresponds to the evaluation of it over the grid points of size $n$. The full data matrix is then obtained by vertically concatenating the different fields (see Fig. \ref{fig: datametrics}b), obtaining a matrix of size $(p \cdot n) \times m$, where $p$ is the number of fields. 

\paragraph{MSFR:} The MSFR dataset comes from a multi-physics numerical simulation of a 2D axisymmetric wedge of the EVOL geometry \cite{brovchenko2013optimization} (see Supplementary Material). The OpenFOAM-based solver couples multi-group neutronic diffusion and incompressible Reynolds-Averaged Navier-Stokes equations \cite{aufiero2014development}, counting more than 20 coupled partial differential equations. In fact, the high-dimensional state space vector $\mathcal{V}$ includes more than 20 different fields, of which we extract five of particular interest to fully describe the dynamics of the system: $q'''_{prompt}, q'''_{decay}, T, u_x, u_z$, being the power density related to prompt and delayed neutrons, the temperature and two velocity components. Among all possible accidental transients, to minimise predictability complex functions were applied to the momentum source (pump velocity) and to the heat sink (heat transfer coefficient and external temperature), starting from steady-state conditions. Albeit not strictly physical, such functions allow observing complex dynamics during the resulting transient, with flow inversions and high-frequency oscillations.

\paragraph{HPMR:} The HPMR dataset is generated from transient simulations of a representative heat pipe microreactor \cite{PRINCE2024110365} that couple time-dependent neutron transport, delayed-neutron precursor evolution, heat conduction, and heat-pipe thermal response.
The transients are reactivity-driven and are initiated by prescribed control-drum motion, which induces changes in both the spatial flux shape and the overall neutron flux magnitude.
These short-time-scale transients are designed to capture the dynamic response of the nuclear system in both prompt- and delayed-neutron-dominated regimes.
The neutron transport solution is obtained using finite-element spatial discretization and discrete ordinates angular discretization \cite{wang2025griffin}, together with an implicit quasistatic acceleration scheme that enables accurate resolution of reactor kinetics over the relevant simulation time scales.
The state space vector $\mathcal{V}$ consists of the reactor power ($P$), component temperatures associated with two fuel zones ($T_{f}$), the hydride moderator temperature ($T_m$), and the graphite monolith temperature ($T_g$).

\paragraph{MHD Channel Flow:} MHD refers to the coupling between magnetic field and fluid-dynamics that occurs in fusion reactors. In particular, the MHD datasets focus on lead-lithium blanket flow within channels \cite{magnetohdfoam2024}: transients are generated by imposing external magnetic fields, varying in magnitude and spatio-temporal shape, which impact the dynamics of the lead-lithium flow and promote magneto-convection depending on the overall magnitude of the magnetic field. The coupled system is solved using finite-volume discretisation and a custom coupled solver. The state space vector consists in the lead-lithium temperature $T$, pressure $p$ and velocity $\mathbf{u}$.

\paragraph{DYNASTY:} The DYNASTY dataset \cite{pini_experimental_2016} is generated using a RELAP5 thermo-hydraulic one-dimensional model of the experimental facility \cite{MISSAGLIA2025107466}. 
DYNASTY was built to study the phenomenology of natural circulation in coupled systems under internal heat generation, akin to the heating and cooling conditions encountered in the MSFR.
The heating transient is driven by imposing different heating conditions to the facility, at different heating power and cooling heat exchanger coefficient parametric values. Additionally, physical thermocouples and a mass flow meter installed in the facility itself are used as additional source of information, providing live experimental data. The dataset temporal focus is on the initial phase of the heating transient before the stabilization of natural circulation. The state space vector consists of temperature $T$ and mass flow rate $\Gamma$.

\paragraph{TRIGA Mark II:} The TRIGA dataset \cite{international2016iaeaTRIGA} comes from a multi-physics (neutronics and thermal-hydraulics) transient simulation of the TRIGA Mark II core, a 250 kW research reactor \cite{TRIGA_MPmodel_yantao}. The transients are initiated by moving the control rods and thus inserting reactivity in the system, changing the shape and magnitude of the neutronic flux and of the coolant temperature. The multi-physics code solves for multi-group neutronic diffusion and three-dimensional thermo-fluid dynamics using finite-volume discretisation and a multi-region approach. The state space vector includes the total flux $\Phi$, the fuel temperature $T_f$, the coolant temperature $T_c$ and the coolant velocity $\mathbf{u}$.

\subsection{Metrics} 

The evaluation metrics of the CTF4Nuclear are designed to comprehensively assess the performance of ML models across a range of tasks and conditions, reflecting the real-world challenges faced in nuclear engineering applications. Following previous works on CTF for scientific ML \cite{wyder2025commontaskframeworkcritical,yermakov2025seismicwavefieldcommontask}, we propose a standard set of metrics for spatio-temporal data, including forecasting, noisy data, limited data and parametric generalisation. Additionally, we introduce novel metrics focused on important aspects for the nuclear engineering community, such as the estimation of the state of the system from sparse sensor measurements, which is a critical aspect in nuclear engineering for monitoring and control purposes \cite{argaud_sensor_2018, cammi_data-driven_2024}. \\
A composite score (\textit{AvgScore}) is then calculated for each dataset by taking the mean of the metrics for each method. Every score is generated by optimizing a single method per task to achieve its highest possible value. All scores are bounded to the range $E_i\in [-100, 100]$. For tasks where a method fails to generate a result, it is assigned the lowest score of -100.

\subsubsection{Standard Set of Metrics}
The 12 scores in the standard set of metrics, taken from previous CTFs \cite{wyder2025commontaskframeworkcritical,yermakov2025seismicwavefieldcommontask}, include\footnote{For more details on the mathematical aspects, the reader is referred to the Supplementary Material and the original papers \cite{wyder2025commontaskframeworkcritical,yermakov2025seismicwavefieldcommontask}.}:
\begin{itemize}
  \item \textbf{Forecasting} (2 scores, E1-E2): evaluate the ability of the model to predict future states of the system based on past observations, with different forecasting horizons (short-term and long-term). 
  \item \textbf{Noisy Data} (4 scores, E3-E6): assess the robustness of the model to noise in the input data, which is common in real-world measurements. This can be done by adding synthetic noise to the test data and evaluating the model's performance. Both reconstruction a.k.a. de-noising (E3, E5) and forecasting (E4-E6) are considered.
  \item \textbf{Limited Data} (4 scores, E7-E10): evaluate the model's performance when trained on limited data, with different forecasting horizons (short-term and long-term) and either noisy or clean data. 
  \item \textbf{Parametric Generalisation} (2 scores, E11-E12): assess the model's ability to generalize to unseen parameter regimes, for interpolation (E11) and extrapolation (E12).
\end{itemize}

\begin{figure}[t]
    \centering
    \includegraphics[width=1\linewidth]{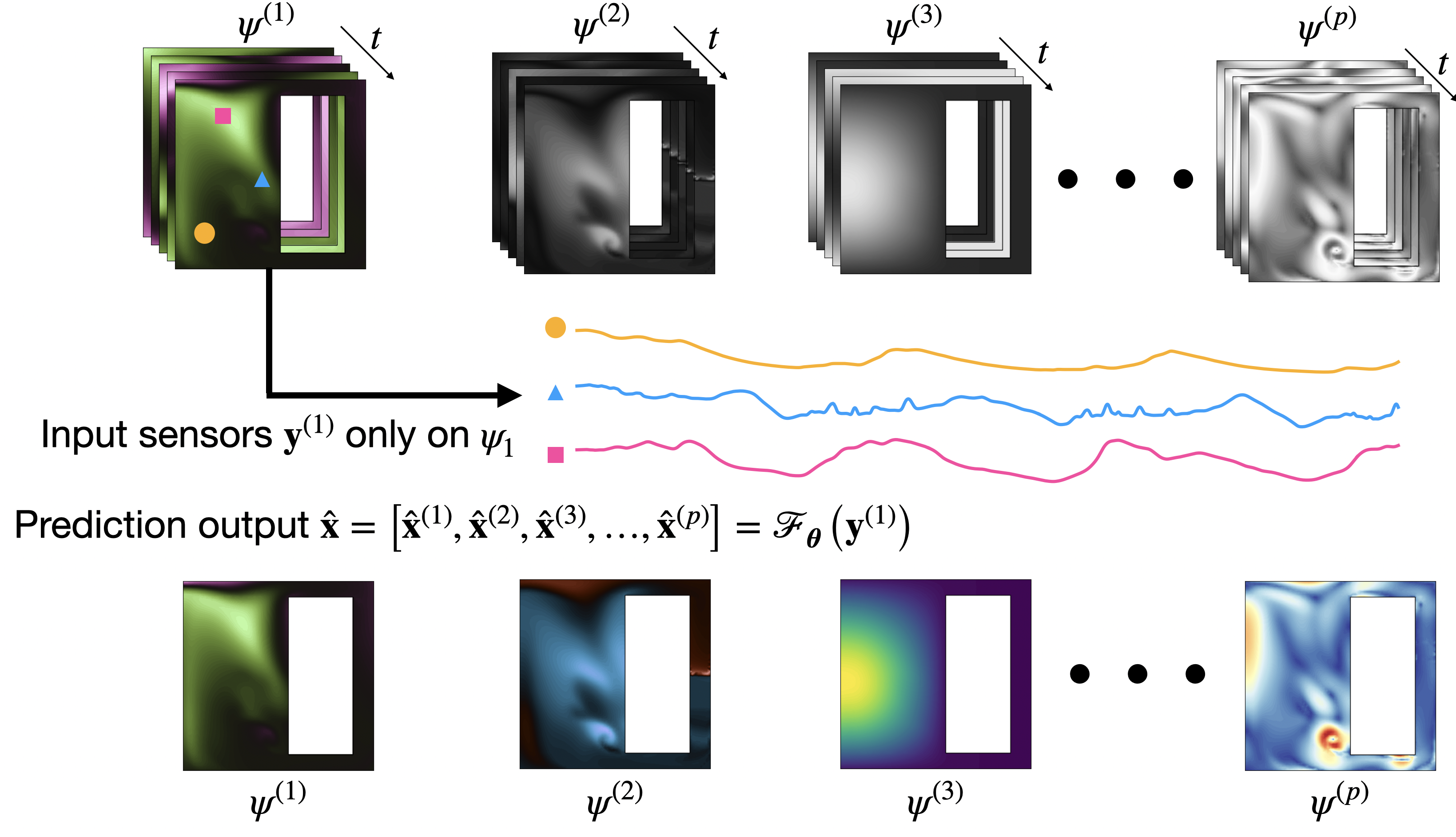}
    \caption{The CTF4Nuclear introduces a new task focused on the reconstruction of the state of the system from sparse sensor measurements. In this task, the model is trained to reconstruct the full state of the system (e.g., neutron flux, temperature, velocity) from a limited number of sensor measurements of a single field.}
    \label{fig: ctf4nuc_newmetrics}
\end{figure}

\subsection{Novel Set of Metrics: Reconstruction from sparse sensors}
In previous works \cite{wyder2025commontaskframeworkcritical,yermakov2025seismicwavefieldcommontask}, it is assumed that the state of the system is fully observable at a certain time $t_j$ to infer the state at a future time $t_{j+1}$. However, in many real-world applications, we might have only access to sparse and partial measurements, in $s$ points, of the state itself. Thus, we may want to change perspective and evaluate the ability of a model to reconstruct the full state of the system from sparse sensor measurements. More specifically, before ML models have been trained using training data (including all the fields) $\mathbf{X}_{\text{train}}$ to predict the test data $\mathbf{X}_{\text{test}}$, adopting the following paradigm $\mathbf{x}_{j+1} = \mathcal{F}_{\boldsymbol{\theta}}(\mathbf{x}_j)$. On the other hand, the novel approach, summarised in Figure \ref{fig: ctf4nuc_newmetrics}, provides as training data the measurements of field $1$ (as an example), $\mathbf{y}^{(1)}_{\text{train}}$, and the full training data matrix $\mathbf{X}_{\text{train}}$, to predict the test matrix $\mathbf{X}_{\text{test}}$ starting from $\mathbf{y}^{(1)}_{\text{test}}$ only. Accordingly, the task is now changed to be $\mathbf{x}_{j} = \mathcal{G}_{\boldsymbol{\theta}}(\mathbf{y}_j^{(1)})$. Since the output is always the same as before, we have not changed how the performance is evaluated thus adopting the same E1-E12 metrics.

{\bf Input:} ${{\bf X}_{\text{train}} = [{\bf X}_{\text{train}}^{(1)},{\bf X}_{\text{train}}^{(2)}, ..., {\bf X}_{\text{train}}^{(p)} ]\in \mathbb{R}^{m \times n}, {\bf y}^{(1)}_{\text{train}}\in \mathbb{R}^{m \times s}},{\bf y}^{(1)}_{\text{test}}\in \mathbb{R}^{m' \times s}, s\ll n$;

{\bf Output:} ${\bf X}_{\text{pred}}=[{\bf X}_{\text{pred}}^{(1)},{\bf X}_{\text{pred}}^{(2)}, ..., {\bf X}_{\text{pred}}^{(p)} ]\in \mathbb{R}^{m' \times n} $.

\section{Methods, Baselines and Results}

In this work, we have investigated several highly-cited modelling methods on the MSFR dataset. Table \ref{full-task-heatmaps} shows all the scored methods and their resulting performance scores on the standard set of metrics of \texttt{ctf4science}. Both regression techniques (e.g., DMD, Koopman-based methods, SINDy), deep learning methods (e.g., LSTM, DeepONet, FNO) and foundation models are included. In addition to the methods, the \textsc{CTF4Nuclear} includes three baselines: a constant zero prediction (Baseline Zeros), a constant average prediction based on the training data (Baseline Averages) and a last-value prediction (Baseline Last), in which the last observed state is used as a prediction for all future states. Table. \ref{full-task-heatmaps} ranks the methods by their average score across all tasks and we notice how only Koopman-based methods are able to outperform the last-value baseline. 
Furthermore, due to the high dimensionality of the spatial domain and the number of fields, some methods have failed to generate a result for some or all the tasks, caused by memory issues/excess of time required to train the model. This is the case for some foundation models such as Moirai-MoE, Chronos-T5, Chronos-2, TabPFN and LLMTime, which have not been designed to handle such high-dimensional spatio-temporal data. The same issue is observed for Spacetime. In order to reduce the dimensionality of the data, we have tried to apply a PCA dimensionality reduction as a pre-processing step, and we were able to obtain results for Reservoir computing and SINDy, but not for the other methods. In the end, we want to recall that if a method fails to generate a result for a task, it is assigned the lowest score of -100.

\begin{table*}[t]
\centering
\caption{Model scores on the MSFR dataset (higher is better; bold marks column maxima), with methods ordered by their average score. Cell colors encode rank within each column (darker = better) using task-specific hues: grayscale for the average, \textcolor{grank3}{\textbf{green}} for forecasting, \textcolor{rrank3}{\textbf{red}} for noisy data, \textcolor{brank3}{\textbf{blue}} for limited data, and \textcolor{yrank3}{\textbf{brown}} for parametric generalisation. We evaluate: Chronos-T5 \citep{ansari2024chronos}, Chronos-2 \citep{ansari2025chronos}, DeepONet \citep{deeponet}, FNO \citep{li2021fourier}, KAN \citep{liu2025kankolmogorovarnoldnetworks}, LLMTime \cite{gruver2023large}, LSTM \citep{hochreiter1997lstm}, Moirai-MoE \citep{liu2024moirai}, Moirai-2 \citep{liu2025moirai}, NeuralODE \citep{ruthotto2024differential}, ODE-LSTM \citep{coelho2024odelstm}, Opt DMD \citep{askham2018variable}, Panda \cite{lai2025panda}, PyKoopman \citep{bruntonkutzkoopmanreview22,Pan2024}, Reservoir \citep{jaeger_echo_no_date, maass_computational_2004, pathak_model-free_2018}, SINDy \citep{sindy, ensemblesindy}, SpaceTime \citep{zhang2023spacetime}, Sundial \citep{liu2025sundialfamilyhighlycapable}, and TabPFN \cite{hoo2025tablestimetabpfnv2outperforms}. Failed runs are assigned a sentinel score of $-100$.}
\label{full-task-heatmaps}
\vspace{4pt}
\scriptsize
\setlength{\tabcolsep}{3pt}
\renewcommand{\arraystretch}{0.95}
\resizebox{\textwidth}{!}{%
\begin{tabular}{l r cc cccc cccc cc}
\toprule
\textbf{Model} & \textbf{Average} & \multicolumn{2}{c}{\textcolor{grank2}{\textbf{Forecasting}}} & \multicolumn{4}{c}{\textcolor{rrank2}{\textbf{Noisy Data}}} & \multicolumn{4}{c}{\textcolor{brank2}{\textbf{Limited Data}}} & \multicolumn{2}{c}{\textcolor{yrank2}{\textbf{Parametric Gen.}}} \\
\cmidrule(lr){3-4} \cmidrule(lr){5-8} \cmidrule(lr){9-12} \cmidrule(lr){13-14}
 & \textbf{Score} & \textbf{E1} & \textbf{E2} & \textbf{E3} & \textbf{E4} & \textbf{E5} & \textbf{E6} & \textbf{E7} & \textbf{E8} & \textbf{E9} & \textbf{E10} & \textbf{E11} & \textbf{E12} \\
\midrule
PyKoopman & \textbf{\pcolortext{1}{70.97}} & \gcolortext{1}{87.46} & \gcolortext{5}{31.12} & \rcolortext{2}{90.91} & \rcolortext{1}{71.81} & \rcolortext{3}{86.92} & \textbf{\rcolortext{1}{85.00}} & \textbf{\bcolortext{1}{92.72}} & \bcolortext{2}{61.21} & \textbf{\bcolortext{1}{92.59}} & \bcolortext{3}{65.95} & \ycolortext{6}{43.36} & \ycolortext{5}{42.56} \\
Baseline Last & \pcolortext{1}{62.67} & \gcolortext{2}{81.20} & \gcolortext{2}{71.79} & \rcolortext{4}{25.45} & \rcolortext{2}{71.79} & \rcolortext{6}{-25.46} & \rcolortext{2}{71.79} & \bcolortext{2}{85.90} & \bcolortext{4}{49.52} & \bcolortext{2}{85.88} & \bcolortext{5}{49.53} & \ycolortext{2}{94.03} & \ycolortext{2}{90.65} \\
Reservoir & \pcolortext{2}{57.22} & \textbf{\gcolortext{1}{92.36}} & \textbf{\gcolortext{1}{87.44}} & \textbf{\rcolortext{1}{98.85}} & \textbf{\rcolortext{1}{88.42}} & \textbf{\rcolortext{1}{98.85}} & \rcolortext{3}{66.02} & \bcolortext{1}{92.09} & \textbf{\bcolortext{1}{92.78}} & \bcolortext{10}{-100} & \bcolortext{10}{-100} & \ycolortext{3}{86.08} & \ycolortext{3}{83.75} \\
SINDy & \pcolortext{2}{55.52} & \gcolortext{3}{75.35} & \gcolortext{4}{42.89} & \rcolortext{3}{41.38} & \rcolortext{3}{42.90} & \rcolortext{3}{41.47} & \rcolortext{4}{42.80} & \bcolortext{4}{73.83} & \bcolortext{3}{52.89} & \bcolortext{2}{73.83} & \bcolortext{3}{52.90} & \ycolortext{3}{70.06} & \ycolortext{4}{55.92} \\
Opt DMD & \pcolortext{3}{53.14} & \gcolortext{4}{72.13} & \gcolortext{4}{41.22} & \rcolortext{2}{90.71} & \rcolortext{4}{41.25} & \rcolortext{2}{91.33} & \rcolortext{5}{41.11} & \bcolortext{4}{60.87} & \bcolortext{4}{50.11} & \bcolortext{3}{61.91} & \bcolortext{4}{52.27} & \ycolortext{6}{36.57} & \ycolortext{7}{-1.80} \\
ODE-LSTM & \pcolortext{3}{52.70} & \gcolortext{3}{80.73} & \gcolortext{3}{61.42} & \rcolortext{5}{5.69} & \rcolortext{4}{30.06} & \rcolortext{2}{93.01} & \rcolortext{4}{52.45} & \bcolortext{3}{78.31} & \bcolortext{5}{42.27} & \bcolortext{5}{34.42} & \bcolortext{2}{66.85} & \ycolortext{5}{56.52} & \ycolortext{5}{30.74} \\
LSTM & \pcolortext{4}{50.07} & \gcolortext{4}{49.26} & \gcolortext{6}{30.35} & \rcolortext{1}{97.02} & \rcolortext{6}{14.21} & \rcolortext{1}{96.39} & \rcolortext{6}{12.15} & \bcolortext{3}{81.42} & \bcolortext{5}{49.22} & \bcolortext{3}{58.03} & \bcolortext{5}{23.59} & \ycolortext{4}{59.12} & \ycolortext{6}{30.02} \\
Moirai-2 & \pcolortext{4}{47.24} & \gcolortext{2}{84.85} & \gcolortext{2}{72.54} & \rcolortext{10}{-100} & \rcolortext{2}{71.72} & \rcolortext{10}{-100} & \rcolortext{2}{66.36} & \bcolortext{2}{89.33} & \bcolortext{3}{53.74} & \bcolortext{1}{88.77} & \bcolortext{4}{51.52} & \ycolortext{1}{94.79} & \textbf{\ycolortext{1}{93.29}} \\
NeuralODE & \pcolortext{5}{40.28} & \gcolortext{5}{37.98} & \gcolortext{3}{60.71} & \rcolortext{6}{-16.67} & \rcolortext{6}{7.20} & \rcolortext{5}{2.24} & \rcolortext{1}{73.27} & \bcolortext{8}{-20.61} & \bcolortext{2}{69.98} & \bcolortext{5}{12.71} & \bcolortext{1}{71.47} & \ycolortext{2}{94.23} & \ycolortext{2}{90.98} \\
DeepONet & \pcolortext{5}{36.64} & \gcolortext{6}{35.42} & \gcolortext{5}{34.42} & \rcolortext{4}{26.79} & \rcolortext{3}{57.75} & \rcolortext{4}{26.23} & \rcolortext{3}{56.80} & \bcolortext{5}{32.97} & \bcolortext{6}{32.33} & \bcolortext{4}{34.90} & \bcolortext{2}{68.81} & \ycolortext{7}{28.22} & \ycolortext{6}{5.02} \\
Sundial & \pcolortext{6}{27.37} & \gcolortext{6}{24.19} & \gcolortext{1}{82.62} & \rcolortext{10}{-100} & \rcolortext{5}{21.82} & \rcolortext{10}{-100} & \rcolortext{5}{27.51} & \bcolortext{6}{9.36} & \bcolortext{1}{86.54} & \bcolortext{6}{6.24} & \textbf{\bcolortext{1}{83.60}} & \textbf{\ycolortext{1}{94.97}} & \ycolortext{1}{91.64} \\
KAN & \pcolortext{6}{5.95} & \gcolortext{7}{9.18} & \gcolortext{7}{0.97} & \rcolortext{3}{72.10} & \rcolortext{5}{21.96} & \rcolortext{4}{3.23} & \rcolortext{7}{0.91} & \bcolortext{6}{5.08} & \bcolortext{7}{-14.10} & \bcolortext{6}{2.31} & \bcolortext{7}{-4.19} & \ycolortext{8}{-7.06} & \ycolortext{8}{-18.96} \\
Baseline Zeros & \pcolortext{7}{0.00} & \gcolortext{8}{0.00} & \gcolortext{7}{0.00} & \rcolortext{6}{0.00} & \rcolortext{7}{0.00} & \rcolortext{6}{0.00} & \rcolortext{7}{0.00} & \bcolortext{7}{0.00} & \bcolortext{7}{0.00} & \bcolortext{7}{0.00} & \bcolortext{6}{0.00} & \ycolortext{7}{0.00} & \ycolortext{7}{0.00} \\
Baseline Averages & \pcolortext{7}{-0.51} & \gcolortext{7}{9.38} & \gcolortext{6}{1.02} & \rcolortext{5}{0.51} & \rcolortext{7}{1.02} & \rcolortext{5}{0.51} & \rcolortext{6}{1.02} & \bcolortext{7}{-14.42} & \bcolortext{6}{5.16} & \bcolortext{7}{-14.42} & \bcolortext{6}{5.16} & \ycolortext{5}{57.27} & \ycolortext{3}{62.69} \\
Panda & \pcolortext{8}{-37.84} & \gcolortext{5}{44.97} & \gcolortext{10}{-100} & \rcolortext{10}{-100} & \rcolortext{10}{-100} & \rcolortext{10}{-100} & \rcolortext{10}{-100} & \bcolortext{5}{32.34} & \bcolortext{10}{-100} & \bcolortext{4}{46.01} & \bcolortext{10}{-100} & \ycolortext{4}{63.99} & \ycolortext{4}{58.57} \\
SpaceTime & \pcolortext{10}{-100} & \gcolortext{10}{-100} & \gcolortext{10}{-100} & \rcolortext{10}{-100} & \rcolortext{10}{-100} & \rcolortext{10}{-100} & \rcolortext{10}{-100} & \bcolortext{10}{-100} & \bcolortext{10}{-100} & \bcolortext{10}{-100} & \bcolortext{10}{-100} & \ycolortext{10}{-100} & \ycolortext{10}{-100} \\
Moirai-MoE & \pcolortext{10}{-100} & \gcolortext{10}{-100} & \gcolortext{10}{-100} & \rcolortext{10}{-100} & \rcolortext{10}{-100} & \rcolortext{10}{-100} & \rcolortext{10}{-100} & \bcolortext{10}{-100} & \bcolortext{10}{-100} & \bcolortext{10}{-100} & \bcolortext{10}{-100} & \ycolortext{10}{-100} & \ycolortext{10}{-100} \\
TabPFN & \pcolortext{10}{-100} & \gcolortext{10}{-100} & \gcolortext{10}{-100} & \rcolortext{10}{-100} & \rcolortext{10}{-100} & \rcolortext{10}{-100} & \rcolortext{10}{-100} & \bcolortext{10}{-100} & \bcolortext{10}{-100} & \bcolortext{10}{-100} & \bcolortext{10}{-100} & \ycolortext{10}{-100} & \ycolortext{10}{-100} \\
Chronos-T5 & \pcolortext{10}{-100} & \gcolortext{10}{-100} & \gcolortext{10}{-100} & \rcolortext{10}{-100} & \rcolortext{10}{-100} & \rcolortext{10}{-100} & \rcolortext{10}{-100} & \bcolortext{10}{-100} & \bcolortext{10}{-100} & \bcolortext{10}{-100} & \bcolortext{10}{-100} & \ycolortext{10}{-100} & \ycolortext{10}{-100} \\
LLMTime & \pcolortext{10}{-100} & \gcolortext{10}{-100} & \gcolortext{10}{-100} & \rcolortext{10}{-100} & \rcolortext{10}{-100} & \rcolortext{10}{-100} & \rcolortext{10}{-100} & \bcolortext{10}{-100} & \bcolortext{10}{-100} & \bcolortext{10}{-100} & \bcolortext{10}{-100} & \ycolortext{10}{-100} & \ycolortext{10}{-100} \\
Chronos-2 & \pcolortext{10}{-100} & \gcolortext{10}{-100} & \gcolortext{10}{-100} & \rcolortext{10}{-100} & \rcolortext{10}{-100} & \rcolortext{10}{-100} & \rcolortext{10}{-100} & \bcolortext{10}{-100} & \bcolortext{10}{-100} & \bcolortext{10}{-100} & \bcolortext{10}{-100} & \ycolortext{10}{-100} & \ycolortext{10}{-100} \\
\bottomrule
\end{tabular}%
}
\end{table*}

\subsection{Observations}

At first, we notice that \textit{PyKoopman} is the only method to 
outperform the last-value baseline, reaching the highest average score over. This is a significant finding, as it suggests that the MSFR transient under study is dominated by low-frequency dynamics, especially since turbulence has been modelled using a Reynolds-Averaged Navier-Stokes approach, which filters out high-frequency fluctuations. The fact that the last-value baseline is so competitive indicates that the system's state changes relatively slowly over time, and that many methods struggle to capture the underlying dynamics beyond this trivial prediction. Furthermore, we can see that regression techniques such as DMD and SINDy are able to perform well over almost all the tasks, being better than deep learning methods such as LSTM, DeepONet and FNO, with Reservoir computing being an exception. This last method performs extremely well for forecasting, noisy data and parametric generalisation, but it fails completely to generalise to limited data. From previous CTFs \cite{wyder2025commontaskframeworkcritical,yermakov2025seismicwavefieldcommontask}, we have noticed that Reservoir computing is a method that can either perform extremely well or fail completely depending on the task, and this is also observed in this work. The fact that regression techniques outperform deep learning methods is also related to the fact that the amount of time series data available for training is relatively small, and the system's dynamics are dominated by a few coherent modes, which favour linear and quasi-linear decompositions over nonlinear architectures. 

The failure of foundation models may be related to a scope mismatch: these architectures are designed for univariate or low-dimensional multivariate time series, and the MSFR dataset comprises $O(10^4)$ spatial degrees of freedom coupled across all physical fields. The only exceptions are Moirai-2, which achieves a respectable average of 47.24, and Sundial which is able to achieve a score of 27.4 adopting a spatial batching strategy. This also illustrates the principle that specialized models often beat general ones in complex domains \cite{goldfeder2026ai}.

\begin{figure}[t]
    \centering
    \includegraphics[width=1\linewidth]{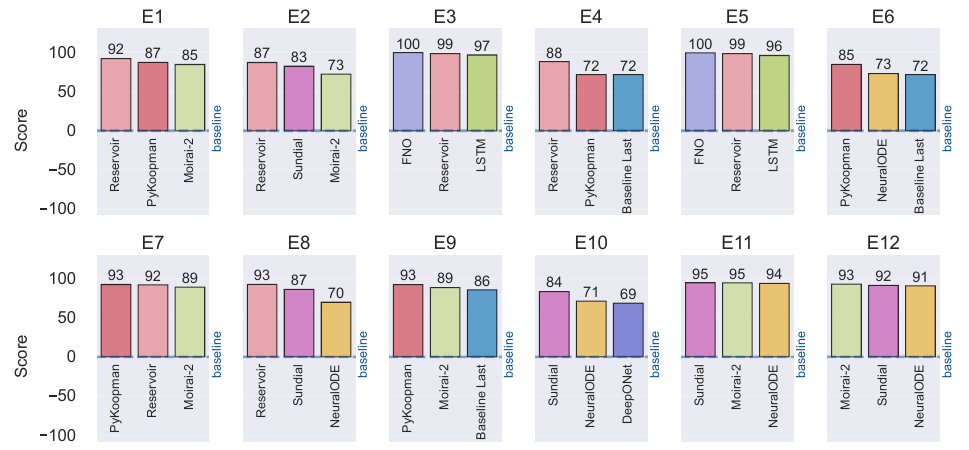}
    \caption{Bar plots of each task from E1 to E12, with the top 3 methods for each task. The blue baseline line corresponds to the constant zero prediction.}
    \label{fig: top3-methods-per-task}
\end{figure}

In the end, Figure \ref{fig: top3-methods-per-task} shows the top three methods for each task, with the blue line corresponding to the constant zero baseline. We can see that PyKoopman, Moirai-2 and Reservoir computing are the most present methods in the top three charts; however, both deep learning methods suffers of a significant drop in specific tasks: for instance, long-term forecasting with limited data (E9 and E10) for Reservoir computing and short-term forecasting with noisy data (E3 and E5) for Moirai-2. On the other hand, PyKoopman is more reliable overall.

\section{Limitations \& Future Work}


We are launching \textsc{CTF4Nuclear} with a first benchmark with the results of several methods on the MSFR dataset, establishing the CTF4Nuclear architecture and baseline methodology. Future iterations and community challenges will expand this repository to include the results of the other datasets described in Section 2.1. In this way, we want to provide a more comprehensive picture of the performance of the different methods across different nuclear systems, with different physics, with different modelling approaches and with different observability constraints. Furthermore, we want to include the new task of reconstruction from sparse sensors, which is a critical aspect for nuclear monitoring and control applications. 

A key limiting factor in achieving high-scores on the current CTF datasets is the small dataset size, which limits the performance of large machine learning models from achieving optimal results. Nevertheless, this was by design and is coherent with the real-world conditions of nuclear engineering applications, where data is often scarce and expensive to obtain. We encourage the machine learning community to develop methods that can perform well under these constraints, and the nuclear community to provide more data either from simulations or from experiments, to establish a well-defined benchmark for scientific machine learning in nuclear engineering.

\section{Conclusions}

The nuclear reactor community (both fission and fusion) lacks a rigorous, standardized framework to evaluate the performance of data-driven surrogate models under the unique multi-physics and observability constraints of nuclear systems. To this end, we introduced \textsc{CTF4Nuclear}, the first comprehensive Common Task Framework specifically tailored to the unique multi-physics and observability constraints of nuclear fission and fusion systems. By incorporating a diversity of datasets (spanning from high-fidelity numerical simulations of molten salt and micro-reactors to real-world experimental facilities) we provide a rigorous proving ground for scientific machine learning. Furthermore, by supplementing standard spatio-temporal metrics with a novel task dedicated to full-state reconstruction from sparse, out-of-core sensors, this framework aligns model evaluation directly with the safety and operational requirements of the nuclear industry.

Our initial benchmarking on the MSFR dataset yields crucial insights into the current state of scientific machine learning. Most notably, standard deep learning architectures and time-series foundation models struggled to process the high spatial dimensionality and low-data regime. Conversely, regression-based approaches and operator theoretic methods, such as PyKoopman or SINDy, demonstrated superior robustness and predictive accuracy, effectively capturing the low-frequency dynamics dominant in the system. These results highlight a critical aspect of modern scientific machine learning: foundational architectures excel in data-rich and low-dimensional domains, but they require significant architectural adaptation to handle the complex, coupled partial differential equations governing reactor physics.

Moving forward, the full release and evaluation of the remaining \textsc{CTF4Nuclear} datasets will provide both the machine learning and nuclear community with a complete view of model generalizability across varying physical phenomena. In the end, we highlight that by exposing the advantages and weaknesses of celebrated learning algorithms we want to stress that the current CTF is not a final verdict on the value of any method; on the contrary, it is an invitation to researchers in the community to refine architectures and to build a more comprehensive benchmark suite for scientific machine learning in nuclear engineering.

\clearpage
\bibliographystyle{plain}
\small{\bibliography{bibliography}}


\appendix


\newpage
\section*{NeurIPS Paper Checklist}

\begin{enumerate}

\item {\bf Claims}
    \item[] Question: Do the main claims made in the abstract and introduction accurately reflect the paper's contributions and scope?
    \item[] Answer: \answerYes{} 
    \item[] Justification: the paper does exactly as stated in the abstract: We build a framework for evaluation scientific machine learning models on nuclear reactor systems, tailoring specific metrics and tasks critical to the industry.

\item {\bf Limitations}
    \item[] Question: Does the paper discuss the limitations of the work performed by the authors?
    \item[] Answer: \answerYes{} 
    \item[] Justification: Yes, we discuss the limitations of our work in a dedicated section. We have highlighted the fact that this work is a first step towards building a comprehensive framework for evaluating scientific machine learning models on nuclear reactor systems.

\item {\bf Theory assumptions and proofs}
    \item[] Question: For each theoretical result, does the paper provide the full set of assumptions and a complete (and correct) proof?
    \item[] Answer: \answerNA{} 
    \item[] Justification: We are benchmarking several models. The assumptions and theoretical results for each model are not applicable for this work, and well beyond the scope of what is attempted to demonstrate here: a fair comparison between methods for scientific machine learning on nuclear reactor systems.  

    \item {\bf Experimental result reproducibility}
    \item[] Question: Does the paper fully disclose all the information needed to reproduce the main experimental results of the paper to the extent that it affects the main claims and/or conclusions of the paper (regardless of whether the code and data are provided or not)?
    \item[] Answer: \answerYes{} 
    \item[] Justification:  The framework is fully available on Github, within the ctf4science Python package to easily replicate all our results, and provide a repository with every evaluated model as a submodule that can be called from the root directory of the main repository. All configuration files used to produce the results are available in the respective model repositories and can be used to reproduce the results.

\item {\bf Open access to data and code}
    \item[] Question: Does the paper provide open access to the data and code, with sufficient instructions to faithfully reproduce the main experimental results, as described in supplemental material?
    \item[] Answer: \answerYes{} 
    \item[] Justification: Reproducibility is the core of this work. Code (\url{https://github.com/CTF-for-Science/ctf4science}) and data (\url{https://osf.io/6rzhm/files/h5qc2}) an extensive appendix are provided to ensure full transparency, access and reproducibility. 

\item {\bf Experimental setting/details}
    \item[] Question: Does the paper specify all the training and test details (e.g., data splits, hyperparameters, how they were chosen, type of optimizer) necessary to understand the results?
    \item[] Answer: \answerNo{} 
    \item[] Justification:  The paper contains all the information on the CTF. Details on the models scored on the benchmark are in the appendix, and the code to reproduce the results on their respective repositories linked above.

\item {\bf Experiment statistical significance}
    \item[] Question: Does the paper report error bars suitably and correctly defined or other appropriate information about the statistical significance of the experiments?
    \item[] Answer: \answerNA{} 
    \item[] Justification: As in previous CTF for scientific ML, their merit as a benchmark doesn't depend on the statistical significance of individual scores and thus error bars were widely omitted. We don't say that there isn't merit to repeated experiment runs or errorbars, but they don't add to or take away from the merit of our work.

\item {\bf Experiments compute resources}
    \item[] Question: For each experiment, does the paper provide sufficient information on the computer resources (type of compute workers, memory, time of execution) needed to reproduce the experiments?
    \item[] Answer: \answerYes{} 
    \item[] Justification: The paper explicitly states that the benchmarking framework and its associated Python package are designed for accessibility and do not require high-performance hardware. The nuclear reactor datasets have been curated to be manageable on average computational resources. If a model requires more compute than what is typically available, we have provided a reduction strategy based on singular value decomposition (See Section 3). 
    
\item {\bf Code of ethics}
    \item[] Question: Does the research conducted in the paper conform, in every respect, with the NeurIPS Code of Ethics \url{https://neurips.cc/public/EthicsGuidelines}?
    \item[] Answer: \answerYes{} 
    \item[] Justification: We ensured full compliance.

\item {\bf Broader impacts}
    \item[] Question: Does the paper discuss both potential positive societal impacts and negative societal impacts of the work performed?
    \item[] Answer: \answerYes{} 
    \item[] Justification: This work is specifically designed to have a positive societal impact by providing a framework for evaluating scientific machine learning models on nuclear reactor systems, which can contribute to safer and more efficient nuclear energy production. However, this might not be restricted to the nuclear community only, we consider this work benign in nature and thus focused our discussion on the groups of machine learning people directly affected by ctf4science.
    
\item {\bf Safeguards}
    \item[] Question: Does the paper describe safeguards that have been put in place for responsible release of data or models that have a high risk for misuse (e.g., pre-trained language models, image generators, or scraped datasets)?
    \item[] Answer: \answerNo{} 
    \item[] Justification: We consider the datasets and framework of ctf4science benign and don't see high-risk for misuse or dual use at this time. The nuclear reactor concepts proposed are specifically designed for energy production only.

\item {\bf Licenses for existing assets}
    \item[] Question: Are the creators or original owners of assets (e.g., code, data, models), used in the paper, properly credited and are the license and terms of use explicitly mentioned and properly respected?
    \item[] Answer: \answerYes{} 
    \item[] Justification: All sources and assets were cited appropriately. 

\item {\bf New assets}
    \item[] Question: Are new assets introduced in the paper well documented and is the documentation provided alongside the assets?
    \item[] Answer: \answerYes{} 
    \item[] Justification: Datasets are documented in the paper and provided in the croissant format. Code is documented and made publicly available. Modelling methods are described in the associated papers and repositories (when applicable).

\item {\bf Crowdsourcing and research with human subjects}
    \item[] Question: For crowdsourcing experiments and research with human subjects, does the paper include the full text of instructions given to participants and screenshots, if applicable, as well as details about compensation (if any)? 
    \item[] Answer: \answerNA{} 
    \item[] Justification: Not applicable, we did not conduct any crowdsourcing experiments or research with human subjects.

\item {\bf Institutional review board (IRB) approvals or equivalent for research with human subjects}
    \item[] Question: Does the paper describe potential risks incurred by study participants, whether such risks were disclosed to the subjects, and whether Institutional Review Board (IRB) approvals (or an equivalent approval/review based on the requirements of your country or institution) were obtained?
    \item[] Answer: \answerNA{} 
    \item[] Justification: Not applicable, we did not conduct any research with human subjects.

\item {\bf Declaration of LLM usage}
    \item[] Question: Does the paper describe the usage of LLMs if it is an important, original, or non-standard component of the core methods in this research? Note that if the LLM is used only for writing, editing, or formatting purposes and does \emph{not} impact the core methodology, scientific rigor, or originality of the research, declaration is not required.
    \item[] Answer: \answerNA{} 
    \item[] Justification: Not applicable.

\end{enumerate}

\end{document}